\documentclass{llncs}

\usepackage{epsfig}
\usepackage{graphicx}
\usepackage{subfig}
\usepackage{makeidx}

\usepackage[utf8]{inputenc}
\usepackage[T1]{fontenc}

\usepackage{amsmath,amssymb, amsfonts} 
\usepackage{gensymb}

\title{Utilization of Deep Reinforcement Learning for saccadic-based object visual search}
\titlerunning{Utilization of Deep Reinforcement Learning for saccadic-based object visual search}

\author{Tomasz Kornuta \and Kamil Rocki}
\authorrunning{T.Kornuta}

\institute{
IBM Research, Almaden, 650 Harry Rd, San Jose, CA 95120, USA\\
\email{\{tkornut,kmrocki\}@us.ibm.com}
}

\def\fig#1{fig.~\ref{fig:#1}}

\def\sec#1{sec.~\ref{sec:#1}}
\def\Sec#1{Sec.~\ref{sec:#1}}

\usepackage{color}

\newcommand{\furl}[1]{\footnote{\url{#1}}}

\begin{document}

\maketitle              

\begin{abstract}
The paper focuses on the problem of learning saccades enabling visual object search.
The developed system combines reinforcement learning with a neural network for learning to predict the possible outcomes of its actions.
We validated the solution in three types of environment consisting of (pseudo)-randomly generated matrices of digits.
The experimental verification is followed by the discussion regarding elements required by systems mimicking the fovea movement and possible further research directions.
\end{abstract}


\section{Introduction}
\label{sec:intro}

\begin{figure}[b!]
	\null\hfill
	\subfloat[]{
		\includegraphics[height=3cm, clip]{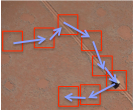}
		\label{fig:saccadic_based_classification}
	}\hfill
	\subfloat[]{\includegraphics[height=3cm, clip]{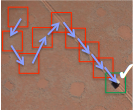}
		\label{fig:saccadic_based_object_detection}
	}\hfill
	\subfloat[]{
		\includegraphics[height=3cm, clip]{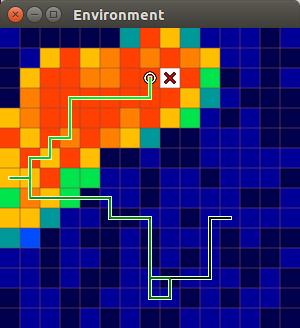}
		\label{fig:saccadic_path}
	}\hfill
	\subfloat[]{
		\includegraphics[height=1.2cm, clip]{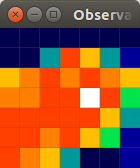}
		\label{fig:saccadic_observation}
	}\hfill\null
	\caption{The desired behaviour of a system realising saccadic-based: (a) image classification/semantic description of the scene, (b) visual object search, (c) an exemplary maze of digits (20x20) with indicated: goal (red cross), current agent pose (white circle) and saccadic path (green line) along with (d) the current agent observation (7x7)}
	\label{fig:model_examples}
\end{figure} 


Humans do not look at a scene in a passive, fixed, steady manner.
Instead, their eyes move around, activelly locating and analysing interesting parts of the scene, constantly building up its mental, three-dimensional model. 
Those rapid jerk-like movements of the eyeball are known as saccades and subserve vision by redirecting the fovea along with the associated visual axis to a new region of interest.
Human eyes fixate mainly on certain elements of the scene that carry or might carry essential or usefull information, whereas saccadic movements depend not only on the objects present in the scene, but also on the task the observer has to achieve~\cite{yarbus1967eye_chapter7}.

There are several possible applications of artificial systems mimicking the fovea movement, with three being particularly interesting, i.e. image classification, semantic description of the scene and visual object search.
For example, as a result of saccadic-based analysis visualized in \fig{saccadic_based_classification} the system might classify the whole image and return a single label ("desert") or return the whole semantic description of the scene (“white truck on a road in a desert”), whereas as a result of the visual object search (\fig{saccadic_based_object_detection}) the system should indicate that the object "white truck" was found in the position (x,y).

In this paper we have focused on the the latter problem, i.e. learning of visual traces mimicking the saccadic motion enabling object search.
\Sec{related_works} briefly presents the general idea and recent applications of Deep Reinforcement Learning, whereas \sec{system-description} describes principles of the operation of the developed system.
In \sec{experimental-verification} we introduce an environment called a maze of digits that we have used for verification of our system along with the obtained results, followed by a brief discussion of the solution and future works in \sec{discussion}.




\section{Related works}
\label{sec:related_works}

Reinforcement Learning (RL), as a general method of injection of the goal and learning goal-oriented behaviurs, had several successfull applications in diverse domains.
For example, as robots possess effectors and receptors enabling them to interact with their environment, RL was the core of a vaist of successfull robotic aplications~\cite{kober2013reinforcement}, including such challenging tasks as 
learning of aerobatic helicopter maneuvers~\cite{abbeel2007application} or
optimization of a humanoid robot gait~\cite{wawrzynski2014reinforcement}.



As Reinforcement Learning in its core relies on the idea of finding action being optimal for a given state, thus the
pure RL-based systems have problems with modelling of huge number of states (or, more generally, with highly dimensional inputs).
For this reason people started to combine RL with Neural Networks (NNs), using the latter as (state) approximators.
Such a combination has a long history. One of the very first examples is TD-Gammon \cite{tesauro1995temporal}, where weights of a neural net were updated according to a  learning rule being a form of temporal-difference (TD) learning. Originally, TD-Gammon stopped improving after about 1,500,000 games (of self-play), reaching a superhuman level.

Recent progress in the end-to-end training of multi-layer neural networks~\cite{krizhevsky2012imagenet} (and in the so called Deep Learning~\cite{lecun2015deep}) once again attracted the attention of researchers to neural nets.
In particular, a lot of attention was put on the mixture of RL with NNs, known recently as Deep Reinforcement Learning (DRL).
In \cite{koutnik2013evolving} authors presented a system combining neural networks with evolutionary programming and reinforcement learning, that was able to learn policies in an end-to-end manner straight from images, enabling it to drive a virtual racing car in a TORCS simulator.
One of the most successful modern examples of such a combination is DQN (Deep-Q-Network) \cite{mnih2015human}. 
The system was based on a Convolutional Neural Network (CNN) trained with RL, reaching super-human level in 29 classic Atari games.
Yet another application of CNN combined with RL (and additionally supplemented with Monte Carlo Tree Search) is AlphaGo \cite{silver2016mastering}, a system the managed to beat both the European (Fan Hui) and World  (Lee Se-dol) Go Champions.
Finally, there are also several interesting works applying Deep Reinforcement Learning in robotics, e.g.
\cite{levine2016end} presented a setup consisting of a battery of 14 robotic manipulators learning simultaneously to grasp different objects and moreover, exchanging gained experience by sharing the policy network, whereas in \cite{gu2016deep} the authors used a similar setup for learning to robustly open doors.


\section{Saccadic-based visual search}
\label{sec:system-description}

\subsection{Deep Reinforcement Learning primer}

Reinforcement  Learning  aims at learning policies controlling actions of an agent interacting with an unknown environment~\cite{sutton1998reinforcement}.
Such an environment is often formalized as a Markov Decision Process (MDP), described  by  a  4-tuple $(\mathcal{S,A,P,R})$.
At each time step $t,$ the agent, being in a state $s_t \in \mathcal{S}$, chooses and executes an action $a_t \in \mathcal{A}$.
In consequence in the next time step the agent receives a reward $r_{t+1} \in \mathcal{R}$ and finds itself in a new state $s_{t+1}$.
$\mathcal{P}$ represents the state transition model of the environment.
The goal of the agent is to maximize the expected discounted reward $R_t = \sum_{k=0}^{\infty} \gamma^k r_{t+k+1}$, with $\gamma$ denoting the discount rate.

Our approach is based on Q-learning~\cite{watkins1989learning}, a Temporal Difference (TD) learning method for estimation of state action values called $Q$-values by updating the current estimate of $Q(s_t, a_t)$ ($Q_t$ in short) towards the reward $R_t$ and estimated utility of the resulting state $s_{t+1}$:
\begin{equation}
\hat{Q}(s_t, a_t) = Q(s_t, a_t) + \alpha (r_t + \gamma (\max\limits_{a'} Q(s_{t+1},a') - Q(s_t, a_t)),
\label{eq:qlearning}
\end{equation}
where $\alpha$ denotes the learning rate.
In particular, when $s_{t+1}$ is the terminal state, then Q-value is simply equal to the final reward associated with that state, i.e. $ \hat{Q}(s_t, a_t) = R(s_{t+1})$.
Such a formulation enables the agent to propagate the rewards being sparsely distributed in the environment from the terminal state to the rest of the environment.

In our system we have decided to combine Reinforcement Learning with a (multi-layer) neural network and use the latter for approximation of Q-values, similarly to DQN~\cite{mnih2015human}.
The NN is parameterized with weights and biases collectively denoted as $\theta_t$.
The input of NN consists of a given state $s_t$, whereas the output of the forward network pass contains the vector of Q-values $Q_t$ associated with all possible types of actions.
This allows to reformulate the TD update rule from \eqref{eq:qlearning} as a rule for update of the parameters of the network realized by minimization of a differentiable loss function:
\begin{equation}
\mathcal{L}(s_t,a_t | \theta_t) \approx (r_t + \gamma (\max\limits_{a'} Q(s_{t+1},a' | \theta_t) - Q(s_t, a_t | \theta_t))^2,
\label{eq:dqn}
\end{equation}
which results in:
\begin{equation}
\theta_{t+1} = \theta_t + \alpha \bigtriangledown_\theta \mathcal{L}(\theta_t).
\end{equation}

Finally, we can also decouple action execution from learning by using Experience Replay.
The idea behind Experience Replay is to collect the experience gained after each action execution, defined as a tuple $e_t =  (s_t, a_t, r_t, s_{t+1})$, and store it in a memory buffer called the Experience Replay memory.
When training, instead of the most recent transition, a random minibatch $\boldsymbol{e}_t$ from the replay memory is used.
The minibatch learning accumulates error (the average error for the whole batch) and performs only a single, aggregated update of network weights calculated on the basis of all samples in a given batch, which significantly improves the convergence.

\subsection{System for learning saccadic-based visual search}
\begin{figure}[htbp]
	\centering
	\includegraphics[width=\textwidth]{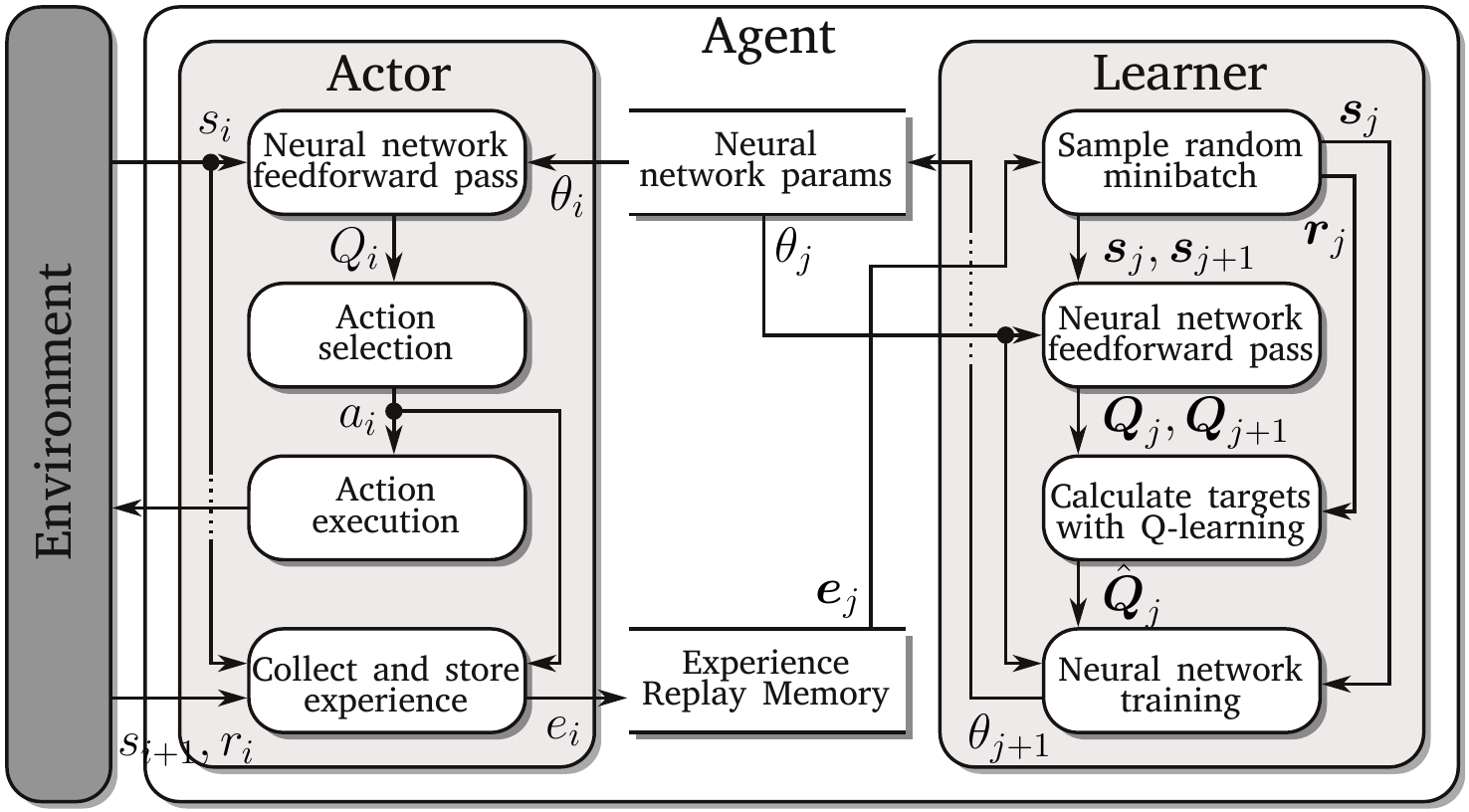}
	\caption{Dataflow diagram of the agent using decoupled DRL}
	\label{fig:dlr_decoupled_acting_learning}
\end{figure}

The data-flow diagram of our system (agent realising visual search) is presented in \fig{dlr_decoupled_acting_learning}.
There are two major components of the system: Actor, responsible for interactions with the environment, and Learner, responsible for learning from the gathered experiences.
Actor and Learner share the parameters of neural network and Experience Replay Memory.

The general principle of the operation of the Actor is as follows.
The current state of the environment $s_i$ (consisting of a single observation, in our case an image patch) in iteration $i$ is passed to the neural network, which predicts the Q-values $Q_i$ received after the execution of four possible types of actions $\mathcal{A} = \{ N,E,S,W \}$, i.e. N (North), E (East), S (South) and W (West).
This enables the actor to decide which action should it perform next (an epsilon-greedy action selection) and execute that action, which results in transition to the next state of the environment $s_{i+1}$.
The actor collects the reward $r_i$, stores the experience $e_i = (s_i, a_i, r_i, s_{i+1})$ in the memory and starts next iteration.

On the other hand, the Learner performs the following operations.
In the first step of a given iteration $j$ it samples a random minibatch $\boldsymbol{e}_j$ from the Experience Replay Memory.
Both vectors of states $\boldsymbol{s}_j$ and $\boldsymbol{s}_{j+1}$ are (independently) passed to the neural network, which results in predictions $\boldsymbol{Q}_j$ and $\boldsymbol{Q}_{j+1}$.
This enables the Learner to calculate the values of $\hat{\boldsymbol{Q}}_j$ according to Q-learning formula \eqref{eq:qlearning}.
It is worth noting that the procedure takes into account only the values for which actions were performed (single value for each experience, as only a single action was performed), whereas other values are simply copied from predictions:

  \begin{equation}
    \hat{Q}(s_j, a_k)=\left\{
                \begin{array}{ll}
                  Q(s_j, a_k) + \alpha (r_j + \gamma (\max\limits_{a'} Q(s_{j+1},a') - Q(s_j, a_k)),& \text{if } a_k  = a_j,\\
                  Q(s_j, a_k),& \text{otherwise},
                \end{array}
              \right.
  \end{equation}
for $a_k$ denoting each one of the four actions possible in a given state.
The resulting values $\hat{\boldsymbol{Q}}_j$ are next used as targets along with the states $\boldsymbol{s}_j$ in training of the neural network using Stochastic Gradient Descent (SGD). 
The resulting parameters $\theta_{j+1}$ are stored in the memory and the next iteration $j+1$ starts.


\section{Experimental verification}
\label{sec:experimental-verification}

\subsection{Partially observable environment: the maze of digits}

The experimental verification of our solution was performed in a controlled environment called "a maze of digits".
Such a maze is simply a matrix of digits (integers) from 0 to 9, with 9 denoting the goal that we want to find or reach.
Those mazes can be threated as simplified, single channel images, where the goal is to find a given characteristic image patch.
In \fig{saccadic_path} the color of the cell indicates the associated digit, the red cross on a white cell denotes the goal, whereas the current position of the agent is represented by a white circle.
The current agent path is plotted with green lines connecting the consecutive grid cells, with brightness indicating the "age" of a given step -- the darker the colour the older the step.
As it is assumed that the system can observe only a part of the environment (\fig{saccadic_observation}), what reformulates the problem as a Partially Observable Markov Decision Process (POMDP). 

\begin{figure}[ht!]
	\null\hfill
	\subfloat[]{
		\includegraphics[height=3cm, clip]{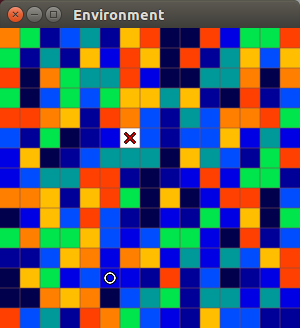}
		\label{fig:random}
	}\hfill
	\subfloat[]{
		\includegraphics[height=3cm, clip]{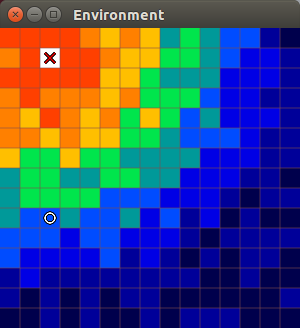}
		\label{fig:circle}
	}\hfill
	\subfloat[]{
		\includegraphics[height=3cm, clip]{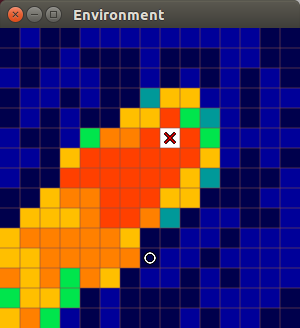}
		\label{fig:path}
	}\hfill\null
	\caption{Different types of "maze of digits": (a) random, (b) circle, (c) path}
	\label{fig:maze_of_digits_elements}
\end{figure}

The experiments were performed with three different types of maze of digits: totally random mazes (\fig{random}), mazes with digits forming a circular structure with the goal located in the center of the structure (\fig{circle}) and mazes with digits forming kind of a path leading to the goal (\fig{path}). The latter type of mazes reflects images containing a car on a road as presented in the introductory example (\fig{saccadic_based_object_detection}), where finding and following one characteristic object in the image (a road) can facilitate search for yet another one (a car).

\subsection{Random mazes of digits}
First we have analysed the convergence in the environment of totally random mazes of digits.
In each experiment we have generated a single maze, whereas at the beginning of every episode we have placed the agent in a different, random position.
As those types of mazes do not possess any clear structure nor tendency indicating how close to the target the agent is, the major goal of those experiments was to check how good the solution is in the memorization of the actions to be performed in different sections of the environment and whether the system can properly learn (approximate Q-values in different states).

\begin{figure}[b!]
	\null\hfill
	\subfloat[]{
		\includegraphics[height=3cm, clip]{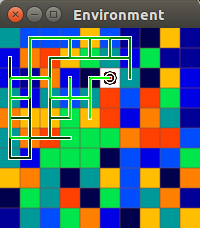}
	}\hfill
	\subfloat[]{
		\includegraphics[height=3cm, clip]{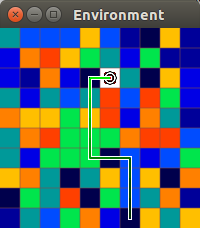}
	}\hfill
	\subfloat[]{
		\includegraphics[height=3cm, clip]{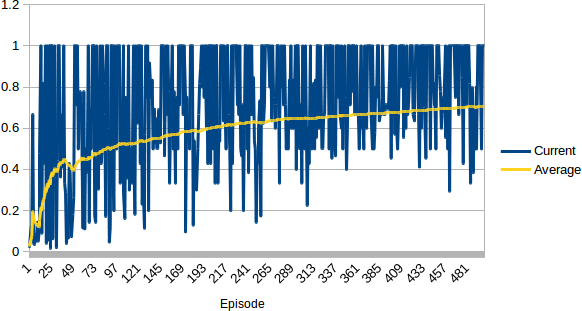}
	}\hfill\null
	\caption{Saccadic paths in random mazes of size 10x10 with observation window 3x3: (a) ep. 0 step 134, (b) ep. 490 step 10, (c) convergence after 500 ep.}
	\label{fig:same_random_10x10_3x3}
\end{figure}

\begin{figure}[ht!]
	\null\hfill
	\subfloat[]{
		\includegraphics[height=3cm, clip]{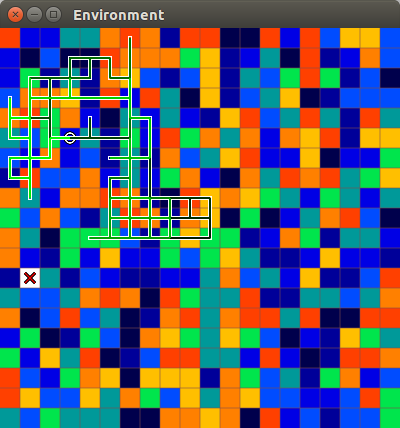}
	}\hfill
	\subfloat[]{
		\includegraphics[height=3cm, clip]{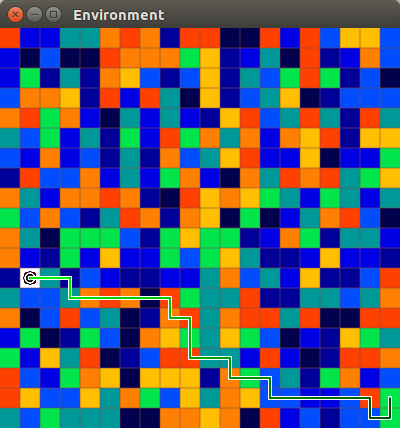}
	}\hfill
	\subfloat[]{
		\includegraphics[height=3cm, clip]{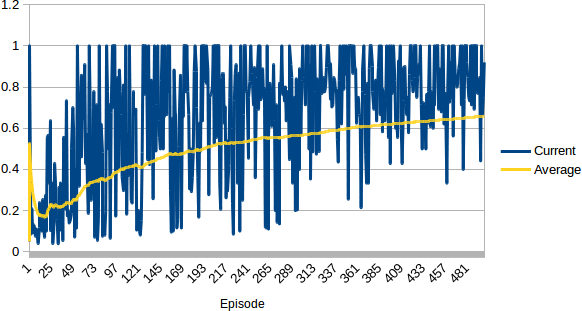}
	}\hfill\null
	\caption{Saccadic paths in random mazes of size 20x20 with observation window 5x5: (a) ep. 0 step 199, (b) ep. 498 step 32, (c) convergence after 500 ep.}
	\label{fig:same_random_20x20_5x5}
\end{figure}


In \fig{same_random_10x10_3x3} and \fig{same_random_20x20_5x5} 
we have presented the exemplary results for mazes of sizes 10x10 and 20x20 and observation windows of sizes 3x3 and 5x5 
respectively.
The plotted "Current" score represents the ratio between optimal (i.e. shortest) path from the initial agent position to the goal and the length of a saccadic path (number of steps), whereas "Average" is the mean running ratio calculated on the basis of all past episodes.
The results prove that the system is able to generate paths leading to the goal for static environments, whereas the convergence depends both on the size of the environment and the size of the window. 
In particular, the bigger the ratio between the size of the observation window and size of the environment, the slower the system was learning.

\subsection{Mazes of digits forming a circle}
Natural images aren’t random -- the majority of our surroundings possess some kind of a structure (forests consist of trees, there are beaches near oceans and seas, houses are surrounded with roads leading to them etc.).
For this reason we have performed a series of experiments with random mazes possessing kind of an underlying structure.
First, we generated mazes with values of digits decreasing along with the growth of distance to the goal. 
Besides that, in the contrary to the previous experiments, in this case in each episode the agent was placed in a newly generated, totally random (thus unique) maze.

\begin{figure}[b!]
	\null\hfill
	\subfloat[]{
		\includegraphics[height=3cm, clip]{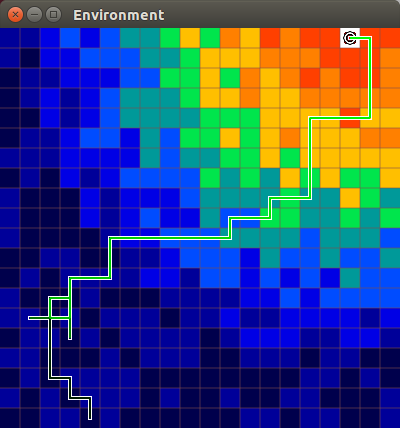}
		\label{fig:circle20x20_episode392_step124_0_26}
	}\hfill
	\subfloat[]{
		\includegraphics[height=3cm, clip]{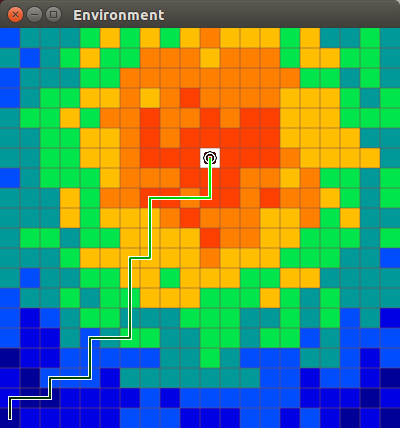}
		\label{fig:circle20x20_episode461_step27_0_85}
	}\hfill
	\subfloat[]{
		\includegraphics[height=3cm, clip]{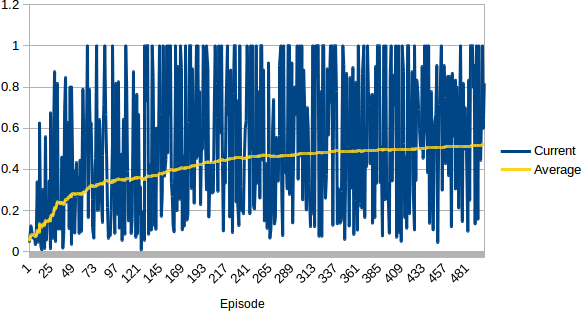}
	}\hfill\null
	\caption{Saccadic paths in circular mazes of size 20x20 and observation window 7x7: (a) ep. 392 step 126, (b) ep. 461 step 27, (c) convergence after 500 ep.}
	\label{fig:circle_20x20_7x7}
\end{figure}

The obtained results (\fig{circle_20x20_7x7}) indicate that the system was able to generalize across different mazes and learn to follow the colour tendency.
The convergence seems slightly worse that in the previous experiments, however it should be stressed out that the utilized metric is very restrictive and even saccades being quite close to the optimal might receive a quite low final score (ratio).
For example, in the episode 392 (\fig{circle20x20_episode392_step124_0_26}) despite that the agent wandered only a bit at the start and managed to succesfully reach the goal -- still, the path was graded the score of 0.26.
Similarly, in \fig{circle20x20_episode461_step27_0_85} the path looks perfect, but along the way the agent made additionally two steps back, which resulted in ratio of 0.85.

\subsection{Mazes of digits forming a path}

Next we validated out system on the random mazes possesing yet another underlying structure, i.e. with digits forming a path leading to the goal.
The experiments were conducted in a similar setup as the ones with random gridworlds, i.e. in each episode a totally new maze was generated and the agent started from a totally different, random position.

\begin{figure}[ht!]
	\null\hfill
	\subfloat[]{
		\includegraphics[height=3cm, clip]{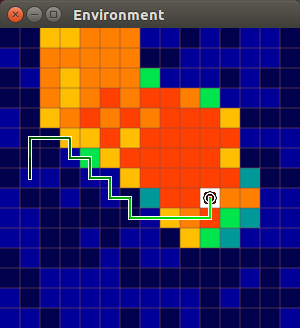}
		\label{fig:path20x20_obs7x7_ep224_step_18_0_56}
	}\hfill
	\subfloat[]{
		\includegraphics[height=3cm, clip]{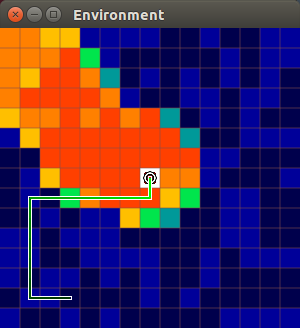}
		\label{fig:path20x20_obs7x7_ep284_step_15_0_73}
	}\hfill
	\subfloat[]{
		\includegraphics[height=3cm, clip]{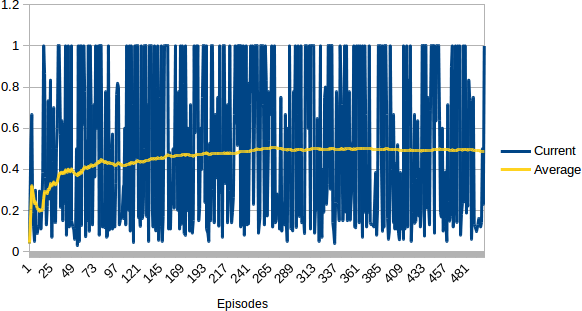}
	}\hfill\null
	\caption{Saccadic paths in circular mazes of size 20x20 and observation window 7x7: (a) ep. 392 step 126, (b) ep. 461 step 27, (c) convergence after 500 ep.}
	\label{fig:road_20x20_7x7}
\end{figure}

Also in this case the system was able to learn the follow that kind of a structural tendency, however the variation of current values is much higher then previously.
This is related to the fact that in many cases the system had to make additional steps, first in order to find the path and then start to follow it.
For example, even though the saccadic paths from \fig{path20x20_obs7x7_ep224_step_18_0_56} and \fig{path20x20_obs7x7_ep284_step_15_0_73} seem quite reasonable, the resulting "Current" scores were 0.56 and 0.76 respectivelly.


\section{Discussion of results and future works}
\label{sec:discussion}

\begin{figure}[t!]
	\null\hfill
		\includegraphics[height=2.8cm, clip]{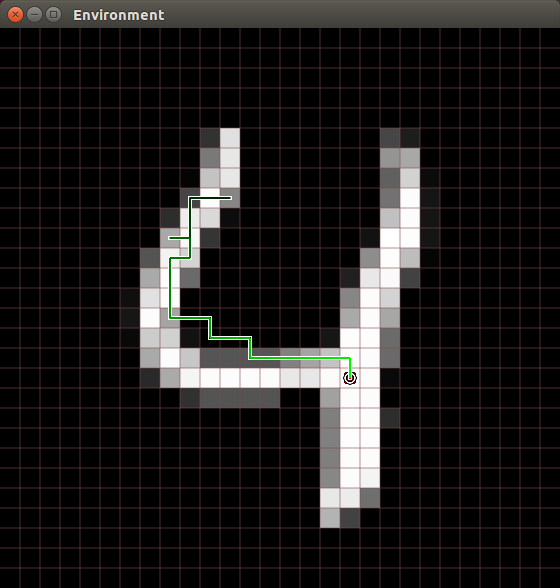}~
		\includegraphics[height=2.8cm, clip]{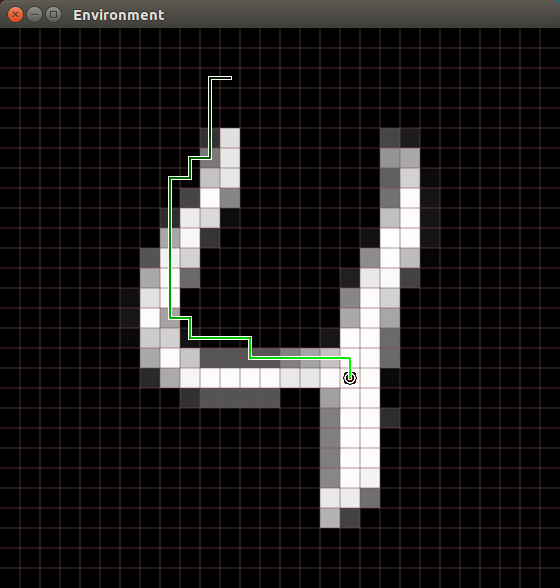}~
		\includegraphics[height=2.8cm, clip]{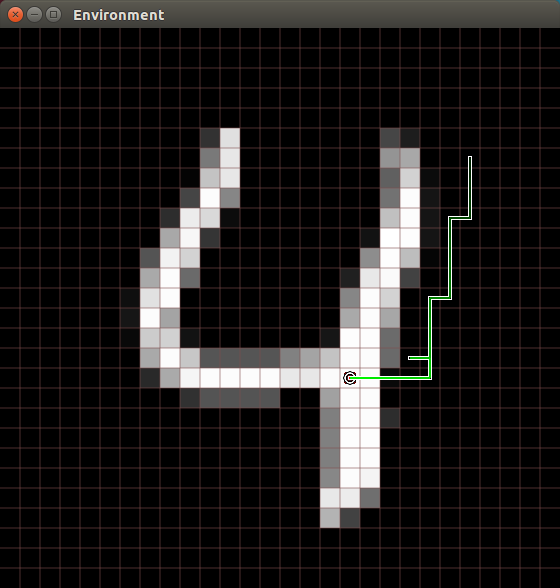}~
		\includegraphics[height=2.8cm, clip]{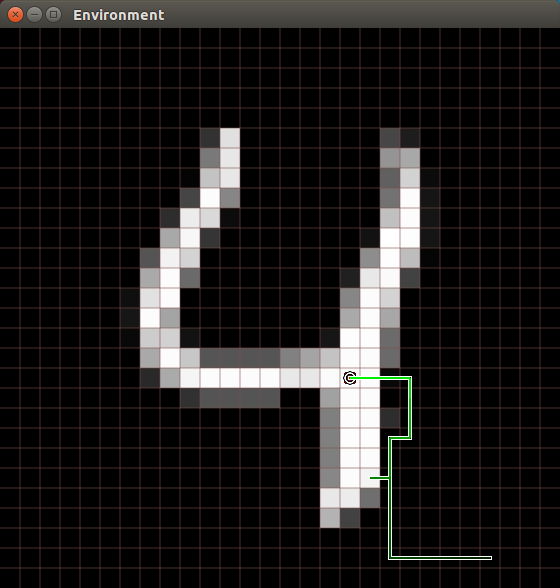}
	\hfill\null
	\caption{Exemplary saccadic paths generated for digit 4}
	\label{fig:mnist_4}
\end{figure}

According to \cite{najemnik2005optimal}, there are three major elements of saccadic system that enable the humans to perform visual search so well, namely:
parallel detection (responsible for finding potential target locations), integration of information accross fixations, and the selection of the next fixation location.
In this paper we have focused on the problem of saccadic visual object search, thus we have narrowed the scope of our research to the third element.
The developed solution, combining reinforcement learning with a neural network, was able to learn in an end-to-end manner and generalize across randomly generated, thus previously unseen environments.
In particular, it was able also to generate saccadic paths matching the shapes of digits from the MNIST dataset (\fig{mnist_4}).
However, in order to get a fully operational saccadic vision system, there are several issues that need to be addressed.

First, the system must be able to aggregate data from consecutive observations.
One possible solution is to use Recurrent Neural Networks (RNNs) such as Long-Short Term Memory (LSTM)~\cite{hochreiter1997long}.
For example, in \cite{le2015simple} the authors successfully used RNNs for classification of MNIST digits represented as a sequence of pixels feed to the network one pixel at a time. 

Next problem concerns more effective analysis and learning of insights regarding the most interesting image fragments and macro-saccadic "jumping" between different parts of the image instead of moving by one pixel in on of four directions. Such a continuous control can be realized in several ways, e.g. in \cite{mnih2014recurrent} the authors used RNN emitting the location of the next image patch to be analysed. 
More effective analysis concerns also the avoidance of already visited places -- one possible solution is to use Neural Turing Machine (NTM)~\cite{graves2014neural}, i.e. RNN with an external memory, for memorization of the already visited locations. 
This also indicates that the system should monitor its current position, e.g. in \cite{mnih2014recurrent} the RNN was feed (aside of multi-scale image patch called glimpse) with a position encoded be a simple NN called location network.

Next, the current solution requires the final rewards to be (manually) placed in the environment, whereas we would like to have the system to autonomously learn rewarding of patches that are more interesting then others.
One possible solution includes training of a network to assign higher value to the patches that are more unique then others (i.e. have smaller inter-patch covariance).
The other possibility is to use RNN to predict the patch associated with the next action/location and use cross-entropy (or surprisal) for grading of that action.
The recent results on utilization of surprisal for learning of long sequences \cite{rocki2016surprisal} prove that it is a good idea.
Those problems and the mentioned possible solutions indicate the future directions of development and research.


\subsubsection*{Acknowledgments}
The authors kindly acknowledge the support of DARPA through the grant "Saccadic Vision and Hierarchical Temporal Memory", contract no. N66001-15-C-4034.

\bibliographystyle{splncs}
\bibliography{2016_automation_dlr_saccades}

\end{document}